\documentclass{ws-procs11x85}
\usepackage{ws-procs-thm}           
\usepackage{multicol}  
\usepackage{multirow}
\usepackage{booktabs}
\usepackage{graphicx}
\usepackage{longtable}
\usepackage{wrapfig}
\usepackage{picinpar}

\DeclareMathOperator*{\argmin}{\arg\!\min}

\begin{document}

\title{Removing Confounding Factors Associated Weights in Deep Neural Networks Improves the Prediction Accuracy for Healthcare Applications}

\author{Haohan Wang$^1$, Zhenglin Wu$^2$, Eric P. Xing$^{1, 3}$}
\address{$^1$School of Computer Science,
Carnegie Mellon University,
Pittsburgh, PA, USA \\
$^2$School of Information Sciences,
University of Illinois Urbana-Champaign
Champaign, IL, USA\\
$^3$Petuum Inc. Pittsburgh, PA, USA\\
E-mail: haohanw@cs.cmu.edu\\
}

\begin{abstract}
The proliferation of healthcare data has brought the opportunities of applying data-driven approaches, such as machine learning methods, to assist diagnosis. Recently, many deep learning methods have been shown with impressive successes in predicting disease status with raw input data. However, the ``black-box'' nature of deep learning and the high-reliability requirement of biomedical applications have created new challenges regarding the existence of confounding factors. In this paper, with a brief argument that inappropriate handling of confounding factors will lead to models' sub-optimal performance in real-world applications, we present an efficient method that can remove the influences of confounding factors such as age or gender to improve the across-cohort prediction accuracy of neural networks. One distinct advantage of our method is that it only requires minimal changes of the baseline model's architecture so that it can be plugged into most of the existing neural networks. We conduct experiments across CT-scan, MRA, and EEG brain wave with convolutional neural networks and LSTM to verify the efficiency of our method.
\end{abstract}



\bodymatter

\section{Introduction}


The increasing amount of data has led healthcare to a new era where the diagnosis can be made directly from raw data such as CT-scan or MRI with data-driven approaches. Machine learning methods, especially deep learning methods, have achieved significant successes in biomedical and healthcare applications, such as classifying lung nodule\cite{hua2015computer}, breast lesions\cite{cheng2016computer}, or brain lesions\cite{gao2017classification} from CT-scans, segmentation of brain regions with MRI\cite{icsin2016review,milletari2017hough}, or emotion classification with EEG data\cite{jirayucharoensak2014eeg,zheng2014eeg}. 

\begin{figure}[h]
\centering
\includegraphics[width=\linewidth]{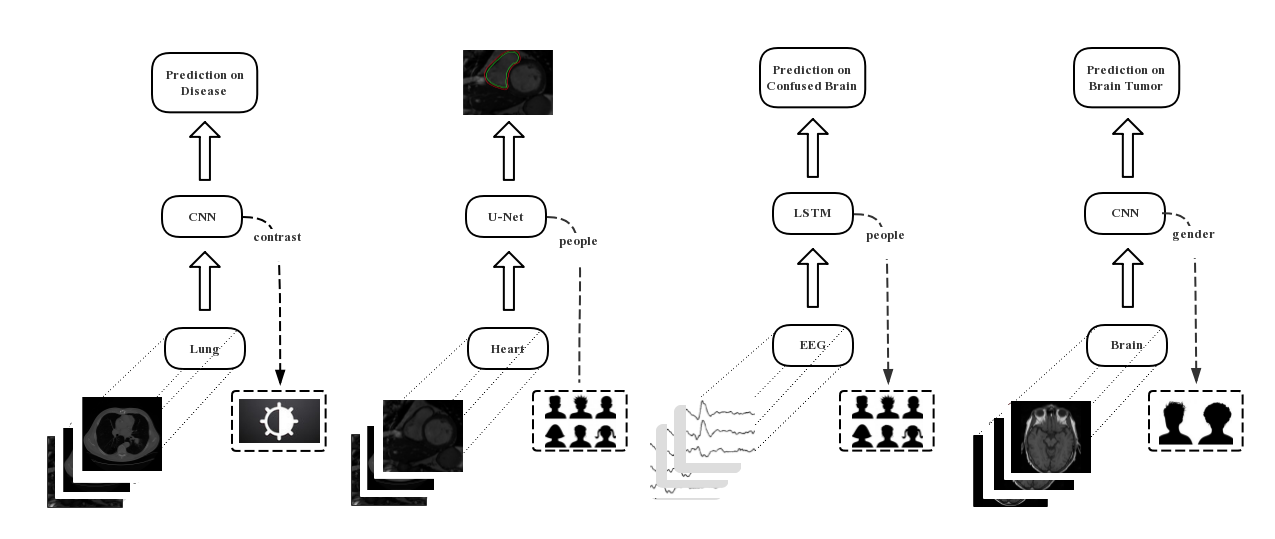}
\caption{An illustration of the empirical contribution of this paper. From left to right, 1) lung adenocarcinoma prediction from CT-scan with CNN, where contrast material is the confounding factor, 2) heart right ventricle segmentation from CT-scan with U-net, where subject identification is the confounding factor, 3) students' confusion status prediction from EEG signals with Bidirectional LSTM, where the students' demographic information is the confounding factor, 4) brain tumor prediction from CT-scan/MRA with CNN, where gender associated information is the confounding factor. }
\label{fig:intro}
\end{figure}

However, different from how deep learning has revolutionized many other applications, the ``black-box'' nature of deep learning and the high-reliability requirement of healthcare industry have created new challenges\cite{miotto2017deep}. One of these challenges is about removing the false signals extracted by deep learning methods due to the existence of confounding factors. Acknowledging the recognition mistakes made by neural networks\cite{szegedy2013intriguing,nguyen2015deep,wang2017origin} and empirical evidence that deep neural networks can learn signals from confounding factors\cite{Wang2017Select}, it is likely that a well-trained deep learning model will exhibit limited predictive performance on external data sets despite its high predictive power on lab collected data sets. The hazard of inappropriate control of confounding factors in healthcare-related science has been discussed extensively\cite{brookhart2010confounding,skelly2012assessing,norgaard2017confounding}, but these discussions are mainly in the scope of causal analyses or association studies. 

In addition to a very recent result showing that confounding factors can adversely affect the predictive performance of neural network models\cite{zech2018confounding}, we offer a straightforward example as another motivation: a neural network predictive model for Hodgkin lymphoma diagnosis is trained on a data set collected from young volunteers with high predictive performance, but when the model is applied to the entire society, it may report more false positives than expected. One of the reason could be that the gender ratio reverses toward adolescence in Hodgkin lymphoma\cite{dorak2012gender}, and a model trained over data collected from young volunteers is very likely to learn a different gender bias than what is expected in a data collected different age groups. In fact, even if the gender ratio does not change along the aging process, it is still inappropriate for a model to predict based on features related to gender because these features are not directly associated with disease status. As another example, skin cancer\cite{syed2012gender} and colorectal cancer\cite{kim2015sex} are also observed with gender bias, and it is already observed that there is a higher false negative rate in colorectal cancer diagnosis for women\cite{kim2015sex} with traditional methods. Confounding factors do not just exist in the forms of gender. Also, it is observed that other factors, such as age\cite{guerreiro2015age}, or demographic information\cite{fincher2004racial}, will affect the model's performance if not handled appropriately. Considering that the generalization theory of neural networks is still an open research topic and people are unsure of how neural networks predict, it is particularly important to design methods to handle the influence of these confounding factors explicitly. 

In this paper, inspired by previous de-confounding techniques applied to deep learning models \cite{Wang2017Select}, we propose a Confounder Filtering (CF) method. A distinct advantage of our method is that CF directly builds upon the original confounded neural network with a minimal change that replaces the original top layer with a layer that predicts the confounding factors. Further, we apply our methods to a broad spectrum of related tasks, such as:
\begin{itemize}
    \item improved lung adenocarcinoma prediction with convolutional neural networks (CNN) by removing contrast material as confounding factors. 
    \item improved heart right ventricle segmentation with U-net by removing subject identifications as confounding factors. 
    \item improved students' confusion status prediction with Bidirectional LSTM by removing students' demographic information as confounding factors. 
    \item improved brain tumor prediction with CNN by removing gender associated information as confounding factors. 
\end{itemize}
We have observed consistent improvements in predictive performance by removing the confounding factors. These four empirical contributions have been conveniently summarized in Figure~\ref{fig:intro}, which illustrates the experiments we perform in this paper, including the predictive task, the model we use, the data, and the confounding factors. 

The remainder of this paper is organized as follows. In Section~\ref{sec:related}, we first briefly discuss the related work of this paper, mainly in the methodological perspective. In Section~\ref{sec:method}, we formally introduce our method, namely Confounder Filtering. Then in Section~\ref{sec:exp}, we apply our method to a wide spectrum of experiments to show the effectiveness of our method and report relevant analysis. Finally, we conclude this paper with discussion of limitations and future directions in Section~\ref{sec:con}. 

\section{Related Work}
\label{sec:related}

The recent boom of deep learning techniques has allowed a large number of neural network methods developed for healthcare applications rapidly. Readers can refer to comprehensive reviews on how the deep learning can be applied to healthcare and biomedical areas\cite{angermueller2016deep,miotto2017deep,ravi2017deep,yue2018deep}. In this section, we will mainly discuss the related work of our paper in the methodological perspective. 

To the best of our knowledge, there are not many deep learning works that control the effects of confounding factors explicitly. Wang \textit{et al} presented a two-phase algorithm named Select-Additive Learning\cite{Wang2017Select}. In the first phase, the model uses information of confounding factors to select which components of the representation learned by neural networks are associated with confounding factors, and then in the addition phase, the algorithm forces the neural networks to discard these components by adding noises. Zhong \textit{et al} also discussed how confounding factors affect the predictive performance of neural networks. They presented an augment
training framework that requires little additional computational costs\cite{zhong2017enlightening}. The idea is to add another neural classifier that predicts confounding factors while predicting original labels, and gradient descent optimizes both of these classifiers. The general additional structure is very similar to the Confounding Filtering method that we are going to present, but our method trains the network in differently so that we can differentiate the weights associated with confounding factors and filter them out explicitly. 

In a broader view, correcting confounding factors is related to reducing the representations learned by neural networks through some components of the raw data that are not related to the predictive task. In this perspective, there is a significant amount of neural network methods that can be considered as related work, covering the fields such as domain adaptation\cite{wang2018deep}, transfer learning\cite{weiss2016survey,moon2014multimodal}, and domain generalization\cite{muandet2013domain}. Readers can refer to the survey papers cited and the references therein if interested. Within the scope of this paper, we do not discuss with these methods for two reasons: 1) these methods are not designed for correcting confounding factors explicitly, therefore they may or may not be applicable in this specific situation, 2) even if our CF method behave similar to, or slightly shy of the performance of these methods, there is still a distinct advantage: CF is simple enough to be plugged into any neural networks with almost no changes of the architecture.

\section{Confounder Filtering (CF) Method}
\label{sec:method}

In this section, we will formally introduce the Confounder Filtering (CF) method. CF method's goal is to reduce the effects of confounders, therefore improves the generalizability of deep neural networks. We first offer an intuitive overview of the main idea of CF, then we formalize our method, which is followed by a discussion of the availability of the implementation.  

\subsection{Overview}

\begin{figure}[h]
\centering
\includegraphics[width=0.8\linewidth]{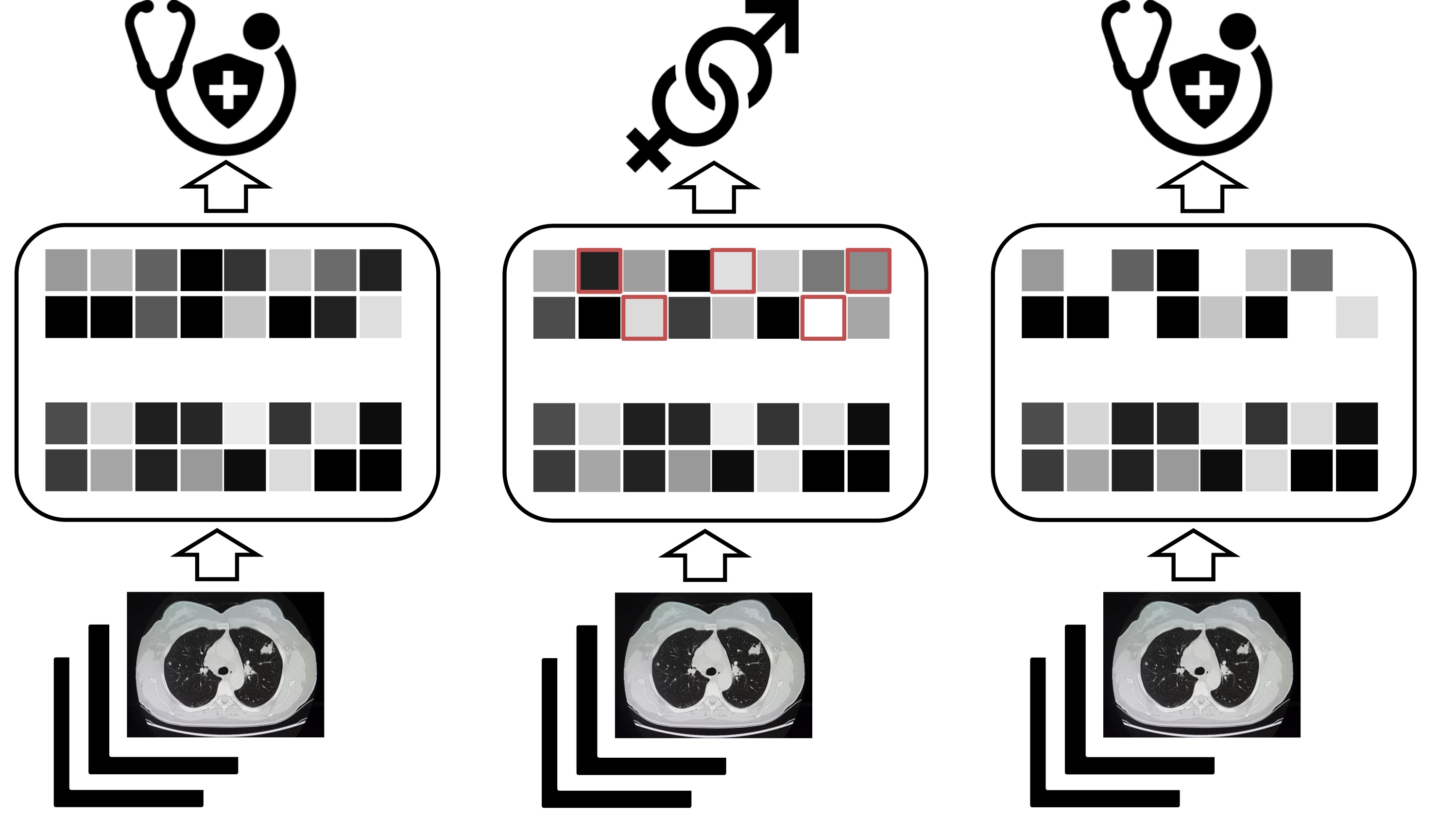}
\caption{This figure shows the overview of the CF method. From left to right: 1) Train the neural network conventionally. 2) Train the neural network to predict confounding factors (\textit{e.g.} gender information) and inspect the changes of weights each iteration to locate the ones with largest changes. 3) Remove the located weights, then the model is ready for confounder-free prediction.}
\label{fig:Model}
\end{figure}

CF method is aimed to remove the effects of confounding factors by removing the weights that are associated with them. Therefore, the core step is to identify such weights. We first train a model, namely $G$, conventionally for the predictive task. Then we replace the top model layer with another classifier that predicts the labels of confounding factors, and we continue to train the model. During this training phase, we keep track of the updates of weights. Finally, we filter out all the weights that are frequently updated during this training phase out of $G$ by replacing these weights with zeros, leading to a new confounder-free model. This process is illustrated in Fig.~\ref{fig:Model}. 

\subsection{Method}
We continue to formalize our method. For the convenience of discussion, we split a deep neural network architecture into two components: representation learner component and classification component, denoted by $g(\cdot ; \theta)$ and $f(\cdot ; \phi)$ respectively, where $\theta$ and $\phi$ stand for the corresponding parameters. Therefore, the complete neural network classifier is denoted as $f(g(\cdot ; \theta);\phi)$. Given data $<y, X>$, the classical training process of the neural network is achieved via solving the following equation:

\begin{align}
\hat{\theta}, \hat{\phi} = \argmin_{\theta, \phi} \; c(y, f(g(X ; \theta);\phi))
\label{eq:nn}
\end{align}

where $c(\cdot, \cdot)$ stands for the cost function, with famous examples such as mean-squared-error loss or cross-entropy loss. 

Ideally, to effectively remove the effects of confounding factors, a method needs the labels of the confounding factors. In other words, we need data in the form of $<X, y, s>$, where $s$ stands for the label of the confounding factors (\textit{e.g.} age, gender, physical factors of medical devices \textit{etc.}). This is also required by similar previous work\cite{Wang2017Select, zhong2017enlightening}. However, our method does not require full correspondence between $X$, $y$, and $s$. For example, later in our experiment, we will show that with two independently collected data sets $<X_1, y_1>$ and $<X_2, s_2>$(\textit{i.e.} we only have correspondence between $X_1$ and $y_1$, and between $X_2$ and $s_2$, but not between $y_1$ and $s_2$), we are able to correct the confounding factors between $X_1$ and $y_1$ with help of $X_2$ and $s_2$. For simplicity, we still present our method with $<X, y, s>$. 

After we train the neural network following the conventional manner as showed in Equation~\ref{eq:nn} with $<X, y>$ and get $\hat{\theta}$ and $\hat{\phi}$, we continue to identify the weights associating with confounding factors through tuning the classification component via $<X, s>$. Formally, we solve the following problem:
\begin{align*}
\tilde{\phi} = \argmin_{\phi} \; c(y, f(g(X ; \hat{\theta});\phi))
\end{align*}

During the optimization, our method inspects how the gradient of the cost function with respect to $<X, s>$ updates the previous trained weights (\textit{i.e.} $\hat{\phi}$) with $<X, y>$. For the $i$\textsuperscript{th} value of $\phi$ (denoted as $\phi_i$), we calculate the frequency of updating it during the entire training process (denoted as $\pi_i$). Formally, we have:

\begin{align*}
\pi_i = \dfrac{1}{n}\sum_{t=1}^n | \Delta \phi_{i,t} |
\end{align*}
where $n$ is the number of total steps, $t$ stands for the index of step, and $I(\cdot)$ is the identify function. 

Further, we construct a masking matrix/tensor $M$ of the same shape as $\phi$, and $M_i$ is constructed according to $\pi_i$. For example, common choices could be either through a Bernoulli sampling
\begin{align*}
M_i = \textnormal{Ber}(\pi_i)
\end{align*}
or a straightforward thresholding procedure:
\[
    M_i=\left\{
                \begin{array}{ll}
                 0, \quad \pi_i > \tau \\
                 1, \quad \textnormal{otherwise}
                \end{array}
              \right.
  \]

In the following experiment, we choose to use the thresholding procedure with $\tau$, whose value lies between top 20\% and top 25\% of $\pi_i$'s values. 

Finally, we have $\hat{\phi}' = \hat{\phi} \otimes M$, where $\otimes$ stands for element-wise product, and the final trained neural network after confounding factor associated weights filtered out is as following:
\begin{align*}
f(g(X ; \hat{\theta}); \hat{\phi}' )
\end{align*}
which is ready for confounder-free prediction.  

\subsection{Availability}
The implementation of our method in TensorFlow is available online\footnote{https://github.com/HaohanWang/CF} with a simple example that trains a CNN for Cifar10 dataset, onto which we add some image patterns as confounding factors. Users can follow the online instruction to apply CF to their own customized neural networks. 

\section{Experiments}
\label{sec:exp}

In this section, we will verify the performance of our CF method on four different tasks by adding CF towards the current baseline models. For each task, we will first introduce the data set, and then introduce the methods we compare and the results. After discussions of these four tasks, we will introduce some analyses of the model behaviors to further validate the performance of our method.  

\subsection{lung adenocarcinoma prediction}
\subsubsection{Data}
We construct a data set to test the model performance in classifying adenocarcinomas and healthy lungs from CT-scans. Our experimental data set is a composition of three data sets: 
\begin{itemize}
    \item \textbf{Data Set 1:} The CT-images from healthy people are collected from ELCAP Public Lung Image Database\footnote{http://www.via.cornell.edu/lungdb.html}. The CT scans have obtained in a single breath hold with a 1.25 mm slice thickness that consists of 1310 DICOM images from 25 persons.
    \item \textbf{Data Set 2:} The CT-scans of diseased lungs are collected from 69 different patients by Grove \textit{et al}\cite{grove2015quantitative}. These scans are diagnostic contrast-enhanced CT scans, being done at diagnosis and prior to surgery and slice thickness at variable from 3 to 6 mm.
    \item \textbf{Data Set 3:} Since these two data sets are collected differently, and one of them is a collection of contrast-enhanced CT scans. The contrast material will likely serve as the confounding factor in prediction. To correct the confounding factor. We noticed a processed version\footnote{https://www.kaggle.com/kmader/siim-medical-images/home} of \textbf{Data Set 2}, which consists of explicit labels of contrast information. The data set contains 475 series from 69 different patients selected 50\% with contrast and 50\% without contrast. 
\end{itemize}

\begin{wrapfigure}{r}{0.5\textwidth}
  \begin{center}
    \includegraphics[width=0.5\textwidth]{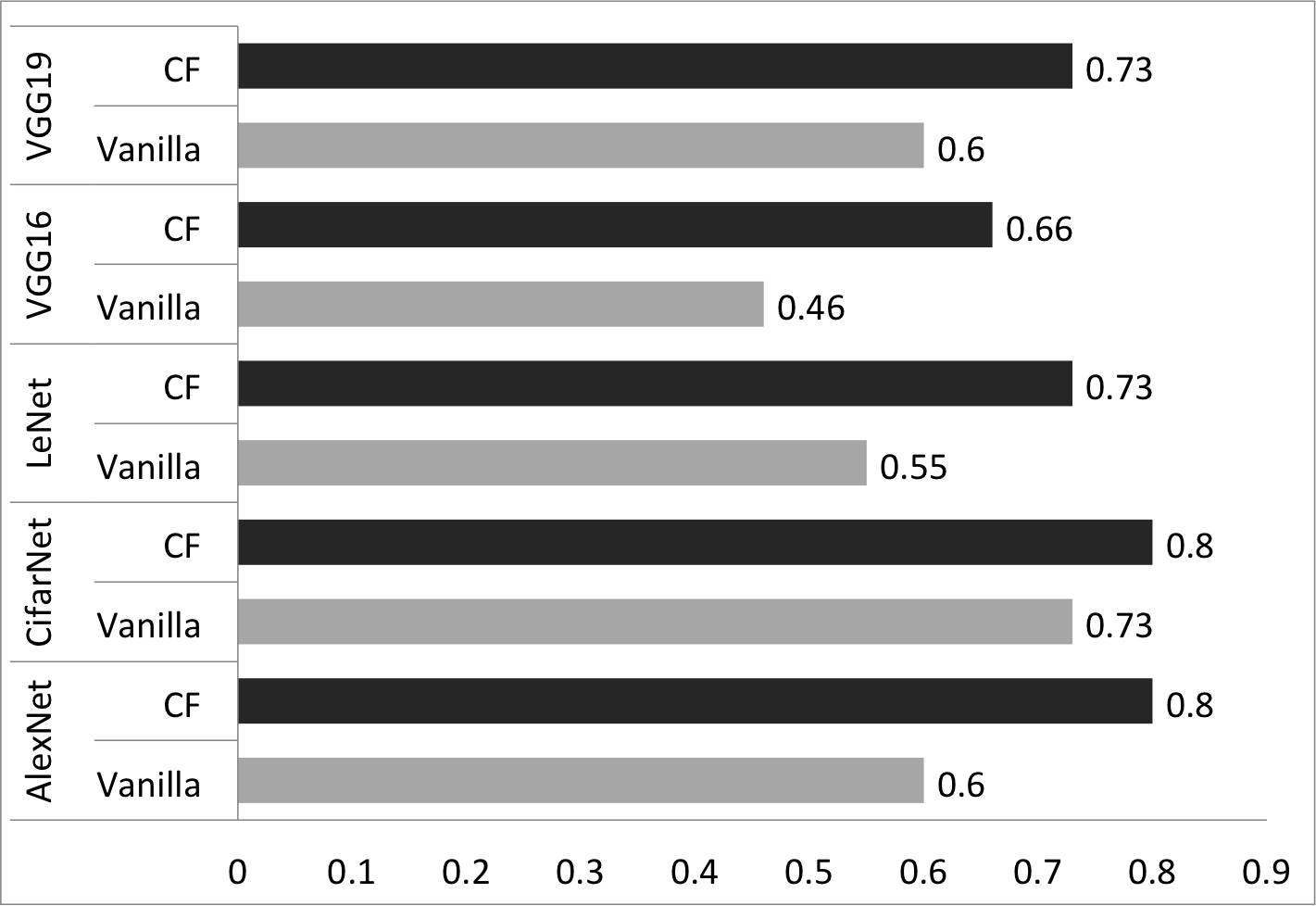}
  \end{center}
  \caption{Prediction accuracy of CNN in comparison with CF-CNN}
  \label{fig:lung}
\end{wrapfigure}

Therefore, we use the 1290 healthy images from 20 persons in \textbf{Data Set 1} and 1214 diseased lung images from 61 patients in \textbf{Data Set 2} as the training set, and the rest from these two data sets as the testing set. We use the images from \textbf{Data Set 3} with corresponding contrast labels to correct confounding factors. 

\subsubsection{Results}

We experiment with the most popular architectures of CNNs, including AlexNet\cite{Krizhevsky2012ImageNet}, CifarNet\cite{Hosang2015Taking}, LeNet\cite{russakovsky2015imagenet}, VGG16\cite{Simonyan2014Very}, and VGG19\cite{Simonyan2014Very}. We first sufficiently train these baseline models, and then use our CF method to correct the confounding factors. We test the prediction accuracy of both vanilla CNNs and CF-improved CNNs. Fig.~\ref{fig:lung} shows the results. We can see that CF can consistently improve the predictive results over a variety of different CNNs.

\subsection{Segmentation on right ventricle(RV) of Heart}

\subsubsection{Data}
The data set \cite{insightdatascience} contains 243 physician-segmented CT images (216$\times$256 pixels) from 16 patients. Data augmentation techniques, such as random rotations, translations, zooms, shears and elastic deformations (locally stretch and compress the image), are used to increase the number of samples. More information regarding the data set, including how the training/testing data sets are split, can be found online\footnote{https://blog.insightdatascience.com/heart-disease-diagnosis-with-deep-learning-c2d92c27e730}. 

\subsubsection{Results}
The main baseline in this experiment is U-net, which is a convolutional network architecture for fast and precise segmentation of images. Previous experiments show that U-net can behave well even with a small dataset\cite{ronneberger2015u}. We first test U-net following previous setting\cite{insightdatascience} and interestingly, we achieve a higher accuracy that what was reported. Vanilla U-net achieves an accuracy of 0.9477. Then, we use CF method to remove the subject identities as confounding factors and improve the accuracy from 0.9477 to \textbf{0.9565}.





\subsection{Students' confusion status prediction}
\subsubsection{Data}
The data set\cite{wang2013using} contains EEG brainwave data from 10 college students while they watch MOOC video clips\footnote{https://www.kaggle.com/wanghaohan/confused-eeg/home}. The EEG data is collected rom MindSet equipment wore by college students when watch ten video clips, five out of which are confusing ones. The students' identities are considered as confounding factors in this experiment. 

\begin{wraptable}{r}{0.5\textwidth}
\normalsize
\centering
\caption{Comparison with average accuracy for 5-fold cross validation \cite{Ni:2017:CCD:3107411.3107513}}
\label{table:eeg}
\begin{tabular}{@{}cccccccccccccccc@{}}
\toprule
\hline
Classification methods       &                   & Accuracy(\%) \\ \hline
SVM                          & \multirow{9}{*}{} & 67.2         \\
K-Nearest Neighbors          &                   & 51.9         \\
Convolutional Neural Network &                   & 64.0         \\
Deep Belief Network          &                   & 52.7         \\
RNN-LSTM                     &                   & 69.0         \\
Bidirectional LSTM           &                   & 73.3         \\
\textbf{CF-Bidirectional LSTM}                 &                   & \textbf{75.0}      \\
\hline\
\end{tabular}
\end{wraptable}

Following previous work\cite{Ni:2017:CCD:3107411.3107513}, we normalize the training data in a feature-wise fashion (\textit{i.e.}, each feature representation is normalized to have a mean of 0 and standard deviation of 1 across each batch of samples). The batch size is set to 20. 

\subsubsection{Results}
We use the state-of-the-art method applied to this data set\cite{Ni:2017:CCD:3107411.3107513}, namely a Bidirectional LSTM, as the baseline method to compare with. The model is configured as following: the LSTM layer has 50 units, with $tanh$ as activation function. The output is connected to a fully connected layer with a sigmoid activation. We compare five-fold-cross-validated results from CF-improved Bidirectional LSTM with results reported previously\cite{Ni:2017:CCD:3107411.3107513}. The results are shown in Table~\ref{table:eeg}. As we can see, CF method helps improve the predictive performance once plugged in. 

\subsection{Brain tumor prediction}
\subsubsection{Data}
We construct another data set for the last experiment of this paper. We test our method in predicting brain tumors with MRA scans of healthy brain\footnote{http://insight-journal.org/midas/community/view/21} and CT-scans with tumor brain\cite{scarpace2015data}. The healthy data set consists of images of the brain from 100 healthy subjects, in which 20 patients were scanned per decade and each group are equally divided by sex. The tumor data set is collected with 120 patients. The gender information is regarded as confounding factors in this experiment. 

Please note that this final experiment has a natural limitation. The healthy brain images are from MRA scans, while the tumorous brain images are from CT scans. The different types of images also serve as a confounder. Therefore, high predictive performance may not may not indicate the corresponding model can be used in industry. However, this experiment is still valid for examining whether CF method can help improve the accuracy relatively. 

\begin{wrapfigure}{r}{0.5\textwidth}
  \begin{center}
    \includegraphics[width=0.5\textwidth]{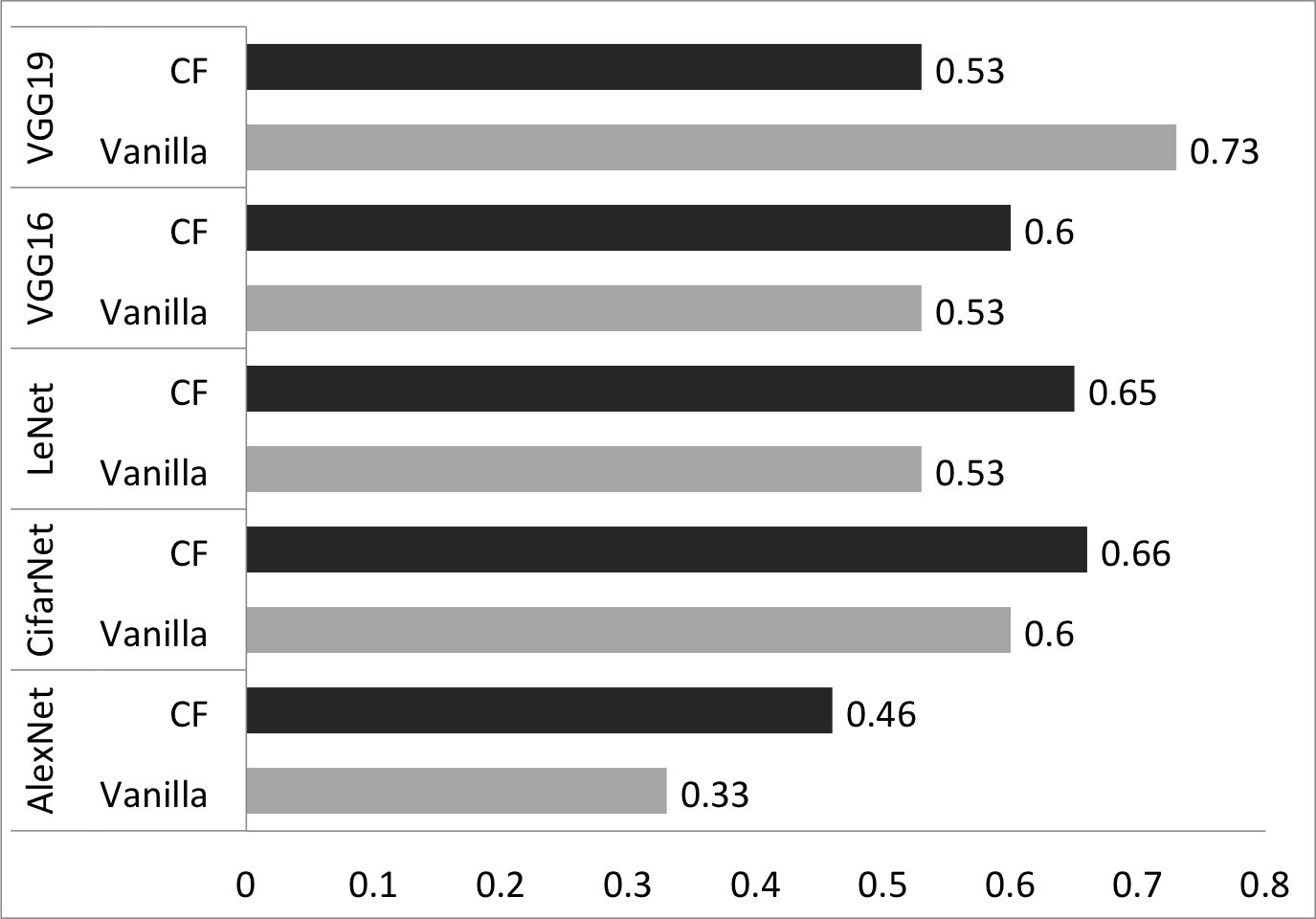}
  \end{center}
  \caption{Prediction accuracy of CNN in comparison with CF-CNN}
  \label{fig:brain}
\end{wrapfigure}

\subsubsection{Results}
Similar to the lung adenocarcinoma prediction experiment, we compare with the set of popular CNNs. The results are shown in Fig.~\ref{fig:brain}. As we can see that, CF helps improve the prediction performance in most cases, except that in the VGG19 cases, when the model's performance deteriorates after CF is plugged in. 

\subsection{Analyses of the method behaviors}
To further understand the process of CF in identifying the weights that are associated with the confounding factors. We inspect how the weights are updated during the training process and visualize which part of the input data is related to confounding factors. 

\begin{figure}[ht]
\centering
\includegraphics[width=\linewidth]{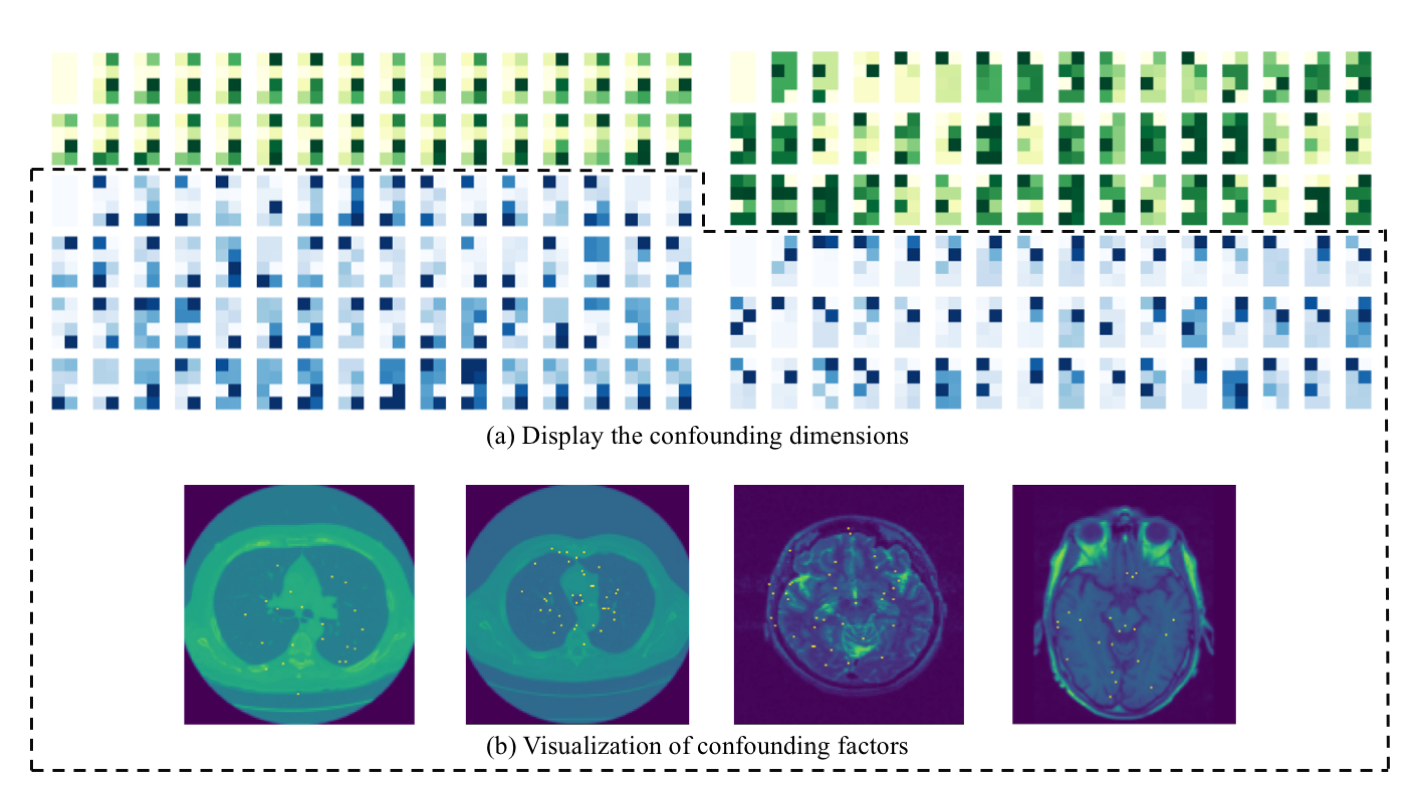}
\caption{(a) Display of trained weights and  (b) the visualization of confounding factors. 
}
\label{fig:lungandbrain}
\end{figure}

Fig.~\ref{fig:lungandbrain}(a) visualizes the weights during each epoch. The figure splits into two panels, and the left panel is for lung adenocarcinoma prediction experiment, and the right panel is for brain tumor prediction experiment. The figure only shows eight weights of the top layer (in a $4\times2$ rectangle), and visualizes how the weights in the layer change as the training epoch increases. This figure visualizes 80 epochs for lung adenocarcinoma prediction and brain tumor prediction each. The blue dots visualize the weights when the model is trained during the first phase, and the green dots visualize the weights when the model is trained in the second phase for prediction confounding factors. The darker each dot is, the more frequent it gets updated in that epoch. As we can see, for the same $4\times2$ layer, the frequencies of the weights get updated are different between the training during the first phase and training during the second phase. This differences of updating frequencies verify the primary assumption of our method, that the weights associated with the task and the weights associated with the confounding factors are different. Therefore, we can remove the effects of confounding factors by removing the weights associated with them. 


Further, we try to investigate which parts of the input data are corresponding to the confounding factors. With the help of Deep Feature Selection\cite{li2015deep} method, we select the pixels of the image that are associated with the confounding factors. Fig~\ref{fig:lungandbrain} visualizes these pixels with yellow dots. From left to right, these four images are examples for healthy lung, diseased lung, healthy brain, tumorous brain respectively. Interestingly, we do not see clear patterns on the images that are related to the confounding factors. This observation further verify the importance of our CF method because these results indicate that it is barely possible to first exclude the information from raw images by conventional methods since these yellow dots do not form into any clear pattern.

\section{Conclusion}
\label{sec:con}

In this paper, we proposed a straightforward method, named Confounder Filtering, which aims to reduce the effects of confounders and improve the generalizability of deep neural networks, to achieve a confounding-factor-free predictive model for healthcare applications. One distinct advantage of our method is that we only require minimal changes to the existing network model to adopt our method. 
We evaluate CF with four healthcare datasets, including CT/MRA of lung, brain, heart and EEG brainwave data. In all these experiments, we have seen consistent improvement of prediction accuracy when the CF method is plugged in. Further, we investigated the process of how the weights are updated during training time to verify the performance of our method. 

A natural limitation of our method is that, despite it only requires a minimal changes of the network architecture, it needs a repeated training process (the second phase training with confounding factors).  In the future, in the methodological perspective, we look forward to further improving the training process of our method. On the practical side, as we have released our code, we hope to help the community to increase the performance of other predictive models for healthcare application by removing the confounding factors. 

\section{Acknowledgement}
The authors would like to thank Mingze Cao and Yin Chen for discussions and creation of Fig~\ref{fig:intro} and Fig~\ref{fig:lungandbrain}. 
This work is funded and supported by the Department of Defense under Contract No. FA8721-05-C-0003 with Carnegie Mellon University for the operation of the Software Engineering Institute, a federally funded research and development center. This work is also supported by the National Institutes of Health grants R01-GM093156 and P30-DA035778.
The MR brain images from healthy volunteers used in this paper were collected and made available by the CASILab at The University of North Carolina at Chapel Hill and were distributed by the MIDAS Data Server at Kitware, Inc

\bibliographystyle{ws-procs11x85}
\bibliography{main}

\end{document}